\documentclass[10pt,twocolumn,letterpaper]{article}

\usepackage[pagenumbers]{cvpr} 
\usepackage{fancyhdr}
\usepackage[dvipsnames]{xcolor}

\definecolor{cvprblue}{rgb}{0.21,0.49,0.74}
\usepackage[pagebackref,breaklinks,colorlinks,citecolor=cvprblue]{hyperref}
\usepackage{lipsum} 

\def\confName{CVPR}
\def\confYear{2024}

\title{Affective Behaviour Analysis via Integrating Multi-Modal Knowledge}

\author{Wei Zhang\textsuperscript{1,$^*$}, Feng Qiu\textsuperscript{1,$^*$}, 
Chen Liu\textsuperscript{1,2}, 
Lincheng Li\textsuperscript{1, $\dagger$}, 
Heming Du\textsuperscript{2}, 
Tiancheng Guo\textsuperscript{2}, 
Xin Yu\textsuperscript{2}\\
\textsuperscript{1} Netease Fuxi AI Lab\\ \textsuperscript{2} The University of Queensland\\
{\tt\small \{zhangwei05, qiufeng, lilincheng\}@corp.netease.com}, 
{\tt\small chen.liu7@uqconnect.edu.au},\\
{\tt\small \{Heming.du, xin.yu\}@uq.edu.au},
{\tt\small alan5gtc@gmail.com}
}

\begin{document}
\maketitle
\def\thefootnote{*}\footnotetext{Equal contribution}
\def\thefootnote{$\dagger$}\footnotetext{Corresponding author}

\begin{abstract}
Affective Behavior Analysis aims to facilitate technology emotionally smart, creating a world where devices can understand and react to our emotions as humans do.
To comprehensively evaluate the authenticity and applicability of emotional behavior analysis techniques in natural environments, 
the 6th competition on Affective Behavior Analysis in-the-wild (ABAW) utilizes the Aff-Wild2, Hume-Vidmimic2, and C-EXPR-DB datasets to set up five competitive tracks, i.e., Valence-Arousal (VA) Estimation, Expression (EXPR) Recognition, Action Unit (AU) Detection, Compound Expression (CE) Recognition, and Emotional Mimicry Intensity (EMI) Estimation.
In this paper, we present our method designs for the five tasks.
Specifically, our design mainly includes three aspects: 1) Utilizing a transformer-based feature fusion module to fully integrate emotional information provided by audio signals, visual images, and transcripts, offering high-quality expression features for the downstream tasks.
2) To achieve high-quality facial feature representations, we employ Masked-Auto Encoder as the visual features extraction model and fine-tune it with our facial dataset.
3) Considering the complexity of the video collection scenes, we conduct a more detailed dataset division based on scene characteristics and train the classifier for each scene.
Extensive experiments demonstrate the superiority of our designs.
\end{abstract}
    
\section{Introduction}
\label{sec:intro}

Affective Behavior Analysis is dedicated to enhancing the emotional intelligence of artificial intelligence systems by analyzing and understanding human emotional behavior \cite{kollias20246th, kollias2020analysing, yin2023multi, kollias2021analysing, zhang2023multi, nguyen2023transformer, ritzhaupt2021meta, kollias2023abaw2, kollias2023abaw, kollias2021distribution, kollias2019expression, kollias2019deep, kollias2019face, zafeiriou2017aff}.
It involves identifying and interpreting the emotions and feelings people express through facial expressions, voice, body language, \emph{etc}.
The goal is to enable computers and robots to better understand human emotional states for more natural and effective human-machine interactions, support mental monitoring, and improve applications in education, entertainment, and social interactions \cite{gervasi2023applications, vsumak2021sensors, filippini2020thermal, szaboova2020emotion, ren2023human}.

The 6th Affective Behavior Analysis competition (ABAW6) has set up the following five tasks to analyze various aspects of human emotions and expressions. 
Action Unit (\textbf{AU}) Detection aims to identify facial action types from the Facia Action Coding System based on facial muscle movements \cite{kollias2023abaw, belharbi2024guided, li2021micro, tallec2022multi}.
Compound Expression (\textbf{CE}) Recognition requires recognizing complex expressions that combine two or more basic expressions \cite{dong2024bi, he2022compound, she2021dive, wang2020suppressing}. 
Emotional Mimicry Intensity (\textbf{EMI}) Estimation evaluates the intensity of an individual's emotional mimicry \cite{wingenbach2020perception, kuang2021effect,  holland2021facial, franz2021your}. 
Expression Recognition (\textbf{EXPR}) identifies basic emotional expressions like happiness, sadness, and anger \cite{li2020deep, revina2021survey, wang2020suppressing, farzaneh2021facial, zhao2021learning}. 
Valence-arousal (\textbf{VA}) estimation determines people's emotional states on continuous emotional dimensions, where ``valence" refers to the positivity or negativity of the emotion, and ``arousal" refers to the level of emotional activation \cite{kollias2021affect, kollias2022abaw, kollias2023abaw, liu2023evaef, praveen2023audio}.

To enhance the applicability of affective behavior analysis techniques in the real world, ABAW6 assesses the method performance on Aff-Wild2 \cite{kollias2018aff}, C-EXPR-DB \cite{kollias2023multi}, and Hume-Vidmimic2 \cite{kollias20246th}, in which videos are captured in uncontrolled natural environments.
Specifically, Aff-Wild2 showcases individuals of different skin tones, ages, and genders, under varied lighting, with assorted backgrounds and head poses, thereby enriching its diversity and applicability.
C-EXPR-DB is designed to analyze multiple emotions that occur simultaneously on the face. 
It consists of videos sourced from YouTube, which feature naturally occurring emotions and expressions.
Hume-Vidmimic2 emphasizes capturing and analyzing the complexity of human emotions in a manner that closely mirrors natural human interactions. 
It bridges the gap between the controlled environment of most emotion recognition datasets and the unpredictability and richness of the natural world.

Based on the characteristics of the above datasets, we establish our objectives to fully utilize the emotional information provided in multimodal data and to enhance the applicability of our method in real-world scenarios.
In this paper, we detail our method designs in three aspects.
Firstly, to obtain high-quality image features. we integrate a large-scale facial image dataset and utilize the self-supervised model Masked Auto Encoder (MAE) \cite{he2022masked, zhang2022transformer} to learn deep feature representations from these emotional data, enhancing the performance of downstream tasks.
Moreover, we leverage a transformer-based model to fuse the multi-modal information.
This architecture facilitates the interactions across modalities (\emph{i.e.,} audio, visual, text) and provides scalable, efficient, and effective solutions for integrating multimodal information \cite{xu2023multimodal}.
Lastly, we adopt an ensemble learning strategy to improve the applicability of our method in various scenes.
In this strategy, we divide the whole dataset into multiple sub-datasets according to their distinct background characteristics and assign these sub-datasets to different classifiers.
After that, we integrate the outputs of these classifiers to obtain the final prediction results.

Experiments conducted on the three datasets demonstrate the effectiveness of our design choices.
Overall, our contributions are three-fold:
\vspace{0.5em}
\begin{itemize}

    \item We integrate a large-scale facial expression dataset and fine-tune MAE on it to obtain an effective facial expression feature extractor, enhancing the performance for downstream tasks.
    \vspace{0.2em}
    \item We employ a transformer-based multi-modal integration model to facilitate the interactions of multi-modalities, enriching the expression features extracted from multi-modal data.
    \vspace{0.2em}
    \item We adopt an ensemble learning strategy, which trains multiple classifiers on sub-datasets with different background characteristics and ensemble the results of these classifiers to attain the final results. This strategy enables our method to generalize better in various environments.
\end{itemize}

\section{Related Work}
\label{sec:relate}

\begin{figure*}[t]
\begin{center}
\includegraphics[width=1.0\linewidth]{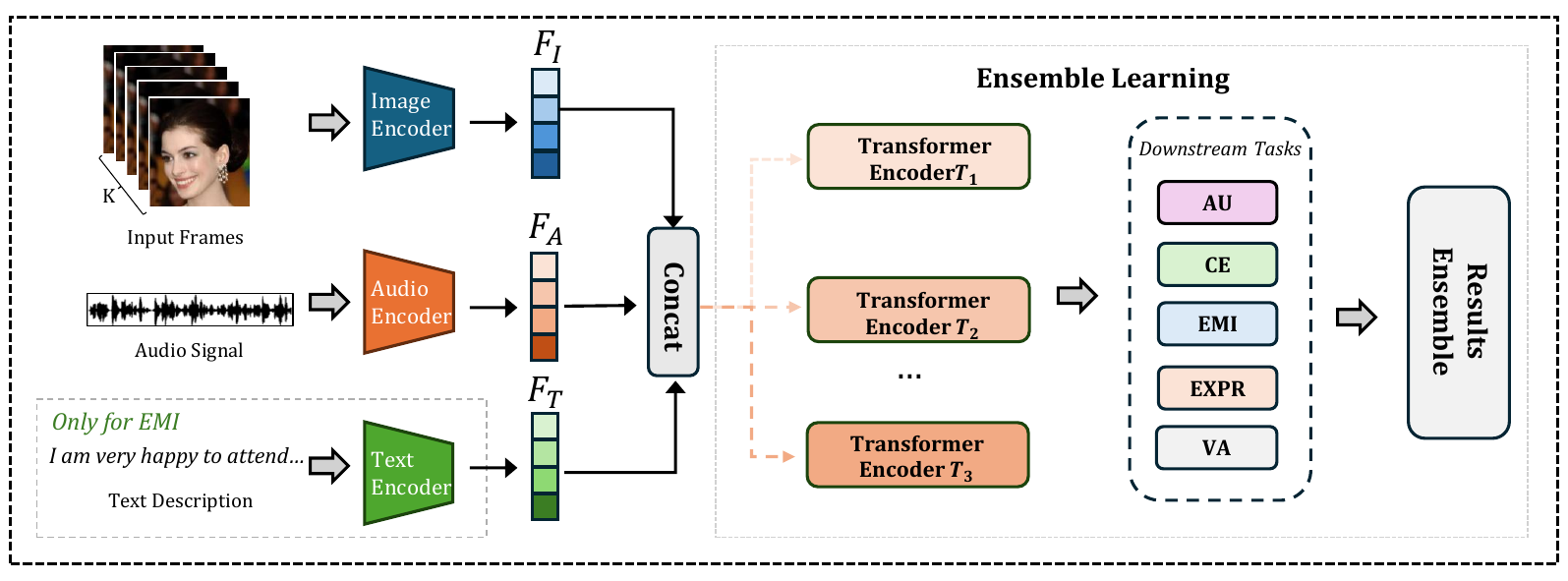}
\end{center}
\vspace{-2em}
\caption{The overview of our proposed method. We first utilize the images in the facial image datasets to train the Image Encoder in a self-supervised manner, thus obtaining the visual feature $F_I$. Then we leverage the pre-trained audio encoder and text encoder to attain the audio feature $F_A$ and text feature $F_T$. Note that we only devise the text encoder for the EMI task. Subsequently, we concat these features and feed them into the Transformer Encoders. Here, we train these encoders on subsets divided based on background characteristics. Finally, we employ a voting strategy to attain the final results.}
\label{fig:pipline}
\vspace{-1.5em}
\end{figure*}

\subsection{Action Unit Detection}
Detecting Action Units (AU) in the wild is a challenging yet crucial advancement task in facial expression analysis, pushing the boundaries of applicability from controlled laboratory settings to real-world environments \cite{li2021micro, tallec2022multi, belharbi2024guided}.
This endeavor addresses the inherent variability in lighting, pose, occlusion, and emotional context encountered in natural environments \cite{kollias2023abaw}.
Recent works highlight the effectiveness of multi-task frameworks in leveraging 
extra regularization, such as the extra label constraint, to enhance detection performance.
Zhang \emph{et al.} \cite{Zhang_2021_ICCV} introduce a streaming model to concurrently execute AU detection, expression, recognition, and Valence-Arousal (VA) regression.
Cui \emph{et al.} \cite{cui2023biomechanics} present a biomechanics-guided AU detection approach to explicitly incorporate facial biomechanics for AU detection.
Moreover, to achieve robust and generalized AU detection, some works take generic knowledge (\emph{i.e.} static spatial muscle relationships) into account \cite{10.5555/3495724.3496926}, while others consider integrating multi-modal knowledge to obtain rich expression features \cite{8578634}.

\subsection{Compound Expression Recognition}
Compound Expression Recognition (CER) gains attention for identifying complex facial expressions that convey a combination of basic emotions, reflecting more nuanced human emotional states \cite{dong2024bi, he2022compound}.
Typical methods focus on recognizing basic emotional expressions with deep learning methods, paving the way for more advanced methods capable of deciphering compound expressions \cite{guo2018dominant, houssein2022human, canal2022survey, yue2019survey}.
Notable efforts in this area include leveraging convolutional neural networks for feature extraction and employing recurrent neural networks or attention mechanisms to capture the subtleties and dynamics of facial expressions over time. Researchers have also explored multi-task learning frameworks to simultaneously recognize basic expressions more accurately and robustly \cite{li2020deep, harbawee2019artificial, wang2023multi}.
Due to the complexity of human emotions in the real world, detecting a single expression is not suitable for real-life scenarios.
Therefore, Dimitrios \cite{kollias2023multi} curates a Multi-Label Compound Expression dataset, C-EXPR. 
Besides, he also proposes C-EXPR-NET, which addresses both CER and AU detection tasks simultaneously, achieving improved results in recognizing multiple expressions \cite{kollias2023multi}.

\subsection{Emotional Mimicry Intensity Estimation}
Emotional Mimicry Intensity (EMI) Estimation delves into the nuanced realm of how individuals replicate and respond to the emotional expressions of others \cite{wingenbach2020perception, kuang2021effect}.
It aims to quantify the degree of mimicry and its emotional impact.
Traditionally, facial mimicry has been quantified through the activation of facial muscles, either measured by electromyography (EMG) or analyzed through the frequency and intensity of facial muscle movements via the Facial Action Coding System (FACS) \cite{ekman1978facial}. 
However, these techniques are either invasive or require significant time and effort.
Recent advancements \cite{dindar2020leaders, dindar2020leaders, varni2017computational} leverage computer vision and statistical methods to estimate facial expressions, postures, and emotions from video recordings, enabling the identification of facial and behavioral mimicry. 
Despite being currently less precise than physiological signal-based measurements, this video-based approach is non-invasive, automatable, and applicable to multimodal contexts, making it scalable for real-time, real-world uses, such as in human-agent social interactions.

\subsection{Expression Recognition}
Expression Recognition has witnessed substantial growth, driven by the integration of psychological insights and advanced deep learning techniques \cite{li2020deep, revina2021survey, wang2020suppressing}.
Recently, the adaptation of transformer-based models from natural language processing (NLP) \cite{vaswani2017attention} to computer vision tasks \cite{dosovitskiy2020image} has led to their application in extracting spatial and temporal features from video sequences for emotion recognition.
Notably, Zhao \emph{et al.} \cite{zhao2021former} introduce a transformer model specifically for dynamic facial expression recognition, the Former-DFER, which includes CSFormer \cite{ye2023csformer} and T-Former \cite{ye2023csformer} modules to learn spatial and temporal features, respectively.
Ma \emph{et al.} \cite{ma2022spatio} developed a Spatio-Temporal Transformer (STT) that captures both spatial and temporal information through a transformer-based encoder. 
Additionally, Li \emph{et al.} \cite{li2022nr} proposed the NR-DFERNet, designed to minimize the influence of noisy frames within video sequences.
While these advancements represent significant progress in addressing the challenges of dynamic facial expression recognition (DFER) with discrete labels, they overlook the interference from the background in images.
To address this, we incorporate ensemble learning into our method.

\subsection{Valence-arousal Estimation}
Valence-arousal estimation focuses on mapping emotional states onto a two-dimensional space, where valence represents the positivity or negativity of emotion, and arousal indicates its intensity or activation level \cite{kollias2021affect, kollias2022abaw, kollias2023abaw}.
Conventional approaches mainly relied on physiological signals, such as heart rate or skin conductance, to estimate these dimensions \cite{bota2019review, lal2023compressed, kranjec2014non}. 
However, with advancements in deep learning, researchers shift towards leveraging visual and auditory cues from facial expressions, voice tones, and body language. 
Notably, convolutional neural networks and recurrent neural networks have been extensively applied to capture the nuanced and dynamic aspects of emotions from images, videos, and audio data \cite{buitelaar2018mixedemotions, yang2018review, marin2020emotion}. 

Recent studies introduce transformer models to better handle the sequential and contextual nature of emotional expressions in multi-modal data \cite{singh2022emoint, chen2020transformer, ju2020transformer}. 
These improvements have not only improved the accuracy and efficiency of valence-arousal estimation but also broadened its applicability in real-world scenarios, such as human-computer interaction and mental health assessment \cite{somarathna2022virtual, dzedzickis2020human, lottridge2011affective}. 
Despite progress, challenges remain in capturing the complex and subjective nature of emotions, necessitating further research into model interpretability and the integration of diverse data sources.

\section{Method}
\label{sec:method}
In this section, we describe our method for analyzing human affective behavior.
The architecture flow is illustrated in Fig. \ref{fig:pipline}.
The proposed approach addresses two critical problems: 1) the emotional information in the multimodal data is not fully explored and 2) the model has poor generalization ability for videos with complex backgrounds. 
For a clear exposition, we first introduce how we utilize the encoders to extract features from multi-modal data in Sec. \ref{sec:encoders}.
Then we detail the transformer-based multi-modal feature fusion method in Sec. \ref{sec:fused}.
Finally, in Sec. \ref{sec:ensemble}, we present the ensemble learning strategy that is leveraged to enhance the model generalization ability.

\subsection{Feature Extraction Encoder}
\label{sec:encoders}
\noindent\textbf{Image Encoder.} 
In this work, we employ MAE as the image encoder since its self-supervised training manner enables the extracted features more generalizable.
To further attain powerful and expressive features, we construct a large-scale facial image dataset which consists of AffectNet \cite{mollahosseini2017affectnet}, CASIA-WebFace \cite{CASIA-Webface}, CelebA \cite{CelebA}, IMDB-WIKI \cite{IMDB-WIKI}, and WebFace260M \cite{zhu2021webface260m}.
The total number of our integrated dataset is 262M.
Based on the integrated facial dataset, we finetune MAE through facial image reconstruction.
Specifically, in the pre-training phase, our method adopts the ``mask-then-reconstruct" strategy.
Here images are dissected into multiple patches (measuring 16$\times$16 pixels), with a random selection of 75\% being obscured.
These masked images are then input into the encoder, while the decoder restores them to the corresponding original.
We adopt the pixel-wise L2 loss to optimize the model, ensuring the reconstructed facial images closely mirror the originals.

After the pre-training, we modify the model for specific downstream tasks by detaching the MAE decoder and incorporating a fully connected layer to the end of the encoder.
This alternation facilitates the model to better adapt to the downstream tasks.

\noindent\textbf{Audio Encoder.}
Considering that the tone and intonation of the speech can also reflect certain emotional information, we leverage VGGish \cite{chen2020vggsound} as our audio encoder to generate the audio representation.
Given that VGGish is trained on the large-scale dataset VGGSound and can capture a wide range of audio features, we directly utilize it as the feature extractor without training on our dataset.

\noindent\textbf{Text Encoder.}
Compared to other tracks, EMI not only provides audio and visual frames but also includes a transcript for each video.
Here, we employ the large off-the-shelf model LoRA \cite{devalal2018lora} to extract features from the transcript.

\subsection{Transformer-based Multi-modal Fusion}
\label{sec:fused}
We fuse features across different modalities to obtain more reliable emotional features and utilize the fused feature for downstream tasks.
By combining information from various modalities such as visual, audio, and text, we achieve a more comprehensive and accurate emotion representation.

To align the three modalities at the temporal dimension, we trim each video into multiple clips with $k$ frames.
For each frame, we employ our image encoder to extract the visual feature $f_I$.
In this fashion, we attain the visual feature $F_{I}^{K \times d}$ for the whole clip.
Here, $d$ represents the feature dimension.
Meanwhile, we employ the audio and text encoders to generate the features for the whole clip, and the features are expressed by $F_{A}^{1 \times d}$ and $F_{T}^{1 \times d}$, respectively.
Subsequently, we concat these features and input them into the Transformer Encoder.
Specifically, our transformer encoder consists of four encoder layers with a dropout rate of 0.3.
The output is then fed into a fully connected layer to adjust the final output dimension according to the task requirements.
Note that, at the feature fused stage, the image encoder, audio encoder, and text encoder are all fixed, while only the transformer encoder as well as the fully connected layer are trainable.

\subsection{Ensemble Learning}
\label{sec:ensemble}
To improve the applicability of affective behavior analysis methods, the 6th ABAW leverages the datasets collected from the real world as the test data.
Given the complex backgrounds in the videos, we adopt the ensemble learning strategy to enable our method robust against complex environments.
Specifically, we first partition the dataset into multiple subsets according to the background characteristics, ensuring each subset contains images with similar background properties.
Next, we separately train the classifiers for each subset to effectively capture emotional information within the images.

During the inference stage, we integrate predictions from classifiers on each subset via a voting method. 
Specifically, for each sample, we allow classifiers from each subset to classify it and record the predictions from each classifier. 
Finally, we employ a voting mechanism based on these predictions to determine the ultimate label.
Here, we select the label with the highest number of votes as the final classification result. 
Our voting method effectively reduces errors caused by biases in classifiers from individual subsets, thereby enhancing overall classification performance.

\subsection{Training Objectives}
\noindent\textbf{Objectives for Image Encoder.}
To enhance the adaptability of the Image Encoder across various tasks, we fine-tune it for each downstream task.
Specifically, when dealing with AU and EXPR, we optimize the model via cross-entropy loss $\mathcal{L}_{AU\_CE}$ and $\mathcal{L}_{EXPR{-}CE}$, respectively. 
They are defined as follows:

\begin{equation}
\mathcal{L}_{AU_{-}CE} = -\frac{1}{12} \sum_{j = 1}^{12} W_{a u_{j}}\left[y_{j} \log \hat{y}_{j}+\left(1-y_{j}\right) \log \left(1-\hat{y}_{j}\right)\right] ,
\end{equation}

\begin{equation}
\mathcal{L}_{EXPR{-}CE } = -\frac{1}{8} \sum_{j = 1}^{8} W_{exp{-}{j}} z_{j} \log \hat{z}_{j},
\end{equation}
where  $\hat{y}$ and $\hat{z}$ represent the predicted results for the action unit and expression category respectively, whereas $y$ and $z$ denote the ground truth values for the action unit and expression category.

In the VA task, to better capture the correlation between valence and arousal and thus improve the accuracy of emotion recognition, we leverage the consistency correlation coefficient as the model optimization function, defined as:
\begin{equation}
\operatorname{CCC}(\mathcal{X}, \hat{\mathcal{X}})=\frac{2 \rho_{\mathcal{X} \hat{\mathcal{X}}} \delta_{\mathcal{X}} \delta_{\hat{\mathcal{X}}}}{\delta_{\mathcal{X}}^{2}+\delta_{\hat{\mathcal{X}}}^{2}+\left(\mu_{\mathcal{X}}-\mu_{\hat{\mathcal{X}}}\right)^{2}},
\label{eq:ccc}
\end{equation}

\begin{equation}
\begin{split}
\mathcal{L}_ {\operatorname{VA}\_\operatorname{CCC}} =1-\operatorname{CCC}(\hat {v}_ {batch_{i}}, v_ {batch_{i}}) \\
+ 1-\operatorname{CCC}(\hat{a}_ {batch_{i}}, a_{batch_{i}}).
\end{split}
\end{equation}

Here, $\hat{v}$ and $\hat{a}$ represent the predicted valence and arousal value. 
$\delta_{\mathcal{X}}$ and $\delta_{\hat{\mathcal{X}}}$ indicate the ground-truth sample set and the predicted sample set.
$\rho_{\mathcal{X} \hat{\mathcal{X}}}$ is the Pearson correlation coefficient between $\mathcal{X}$ and $\hat{\mathcal{X}}$, 
$\delta_{\mathcal{X}}$ and $\delta_{\hat{\mathcal{X}}}$ are the standard deviations of $\mathcal{X}$ and $\hat{\mathcal{X}}$, and $\mu_{\mathcal{X}}$, $\mu_{\hat{\mathcal{X}}}$ are the corresponding means. The numerator $2 \rho_{\mathcal{X} \hat{\mathcal{X}}} \delta_{\mathcal{X}} \delta_{\hat{\mathcal{X}}}$ represents the covariance between the $\delta_{\mathcal{X}}$ and $\delta_{\hat{\mathcal{X}}}$ sample sets.

\noindent\textbf{Objectives for Transformer-based Fusion Model.}
For each task, we utilize the same training objective as the Image Encoder for optimizing the transformer-based fusion model.
Additionally, since the generated results are frame-wise rather than at the clip level, we employ a smoothing strategy to improve the consistency of the predictive results.

Specifically, our strategy is conducted in two steps, we first utilize the facial detection model RetinaNet \cite{deng2020retinaface} to identify which frame suffers from face loss due to the crop data augmentation operation and then replace the frames without faces with adjacent frames to ensure the integrity of faces in the sequence. 
In the second step, we leverage a Gaussian filter to refine the likelihood estimations for AU,  EXPR, as well as VA.
It is formulated as:
\begin{equation}
\mathcal{L}_{smooth} = \int_{-\infty}^{\infty} \left( y - \frac{\sqrt{2} \cdot f(x) \cdot e^{-\frac{(x - \mu)^2}{2\sigma^2}}}{2\sqrt{\pi}\sigma} \right)^2 dx,
\end{equation}
where $y$ represents the predicted value of the five downstream tasks, $f(x)$ is the predicted likelihood estimation before applying the Gaussian filter, and $e$ is the base of the natural logarithm.
$x$ and $\mu$ represent the input value and the mean of the distribution, respectively.
$\sigma$ indicates the standard deviation of the distribution, determining the width of the Gaussian curve.
The Gaussian filter's sigma parameter is tuned specifically for each task, with the precise configurations detailed in the experimental setup section.

\section{Experiment}
\label{sec:exp}
In this section, we will first introduce the evaluation metrics datasets as well as the implementation details. Then we evaluate our model on the ABAW6 competition metrics.

\subsection{Evaluation metrics}
To assess the model performance on each track, ABAW set a specific evaluation metric for each track.

\noindent\textbf{Valence-Arousal Estimation.} The performance measure (P) is the mean Concordance Correlation Coefficient (CCC) of valence and arousal, as follows:
\begin{equation}
    P = \frac{\operatorname{CCC}_{arousal} + \operatorname{CCC}_{valence}}{2}.
\end{equation}
Here, the calculation method of $\operatorname{CCC}$ is defined in Eq. \ref{eq:ccc}.

\noindent\textbf{Expression Recognition.} The performance assessment is conducted by averaging F1 score across all 8 categories, defined as:
\begin{equation}
\left\{
\begin{aligned}
& F1 = \frac{2 \times {Precision} \times {Recall}}{{Precision} + {Recall}}; \\
& {Precision} = \frac{{TP}}{{TP} + {FP}}; \\
& {Recall} = \frac{{TP}}{{TP} + {FN}},
\end{aligned}
\right.
\label{eq:F1}
\end{equation}

\begin{equation}
    P = \frac{\sum_{c=1}^8F1_c}{8}.
\end{equation}
Here, $c$ represents the category ID, $TP$ represents True Positives, $FP$ represents False Positives, and $FN$ represents False Negatives.

\noindent\textbf{Action Unit Detection}. The performance is evaluated by averaging the F1 score across all 12 categories, formulated as:
\begin{equation}
    P = \frac{\sum_{c=1}^{12} F1_{c}}{12}
\end{equation}
Here, the calculation way of $F1$ is the same as the Eq. \ref{eq:F1}.

\noindent\textbf{Compound Expression Recognition.} In this track, the performance measure P is the average F1 Score across all 7 categories, calculated as:
\begin{equation}
    P = \frac{\sum_{c=1}^{7} F1_{c}}{7}
\end{equation}
Here, the calculation way of $F1$ is the same as the Eq. \ref{eq:F1}.

\noindent\textbf{Emotional Mimicry Intensity Estimation.}
EMI evaluates the performance by averaging Pearson's correlations ($\rho$) across the emotion dimensions, defined as:
\begin{equation}
    P = \frac{\sum_{c=1}^6\rho_{c}}{6}.
\end{equation}

\begin{table*}[!t]
\small
\renewcommand\arraystretch{1.05} 
\centering
  \caption{The AU F1 scores~(in \%) of models that are trained and tested on different folds (including the original training/validation set of \textit{Aff-Wild2} dataset).}
  \label{tab:AU_F1_val}
    \vspace{-0.7em}

\setlength{\tabcolsep}{0.6 em}
\renewcommand\arraystretch{1.2}
  \begin{tabular}{l|cccccccccccc|c}
    \hline
    \textbf{Val Set} &\textbf{AU1} &\textbf{AU2} &\textbf{AU4} &\textbf{AU6} &\textbf{AU7} &\textbf{AU10} &\textbf{AU12} &\textbf{AU15} &\textbf{AU23} &\textbf{AU24} &\textbf{AU25} &\textbf{AU26} &\textbf{Avg.}\\[1pt]
    \hline \hline
    Official & 55.29 & 51.40 & 65.81 & 68.61 & 76.08 & 75.00 & 75.24 & 37.65 & 18.89 & 30.89 & 83.41 & 44.98 & 56.94\\
    fold-1   & 62.61 & 46.20 & 71.22 & 77.71 & 67.44 & 69.69 & 74.62 & 36.32 & 29.43 & 21.75 & 81.56 & 40.73 & 56.61 \\
    fold-2   & 64.23 &  54.35 & 73.85 & 77.33 & 77.49 & 76.70 & 80.74 & 29.05 & 28.96 &  18.47& 87.71 & 43.63 & 59.37\\
    fold-3   & 58.55 & 48.37 & 60.05 & 71.22  & 72.43 & 74.29 & 75.43 & 29.81 & 19.52 & 32.86 & 83.37 & 47.63 & 56.13\\
    fold-4   & 53.34 & 39.34 & 66.26 & 70.67 & 66.51 & 69.39 & 71.76 & 39.49 & 25.17 & 32.40 & 82.27 & 40.05 & 54.72\\
    fold-5   & 53.50 & 44.68 & 63.45 & 72.02 & 69.72 & 74.00 & 78.24 & 38.81 & 23.67 & 7.56 & 81.24 & 43.67 & 54.22\\
    \hline
  \end{tabular}
\label{exp:au}
\end{table*}
\subsection{Datasets}
The first tracks of ABAW6 are based on Aff-wild2 which contains around 600 videos annotated with AU, base expression category, and VA. 
The AU detection track utilizes 547 videos of around 2.7M frames that are annotated in terms of 12 action units, namely AU1, AU2, AU4, AU6, AU7, AU10, AU12, AU15, AU23, AU24, AU25, AU26. The performance measure is the average F1 Score across all 12 categories. 
The expression recognition track utilizes 548 videos of around 2.7M frames that are annotated in terms of the 6 basic expressions (i.e., anger, disgust, fear, happiness, sadness, surprise), plus the neutral state, plus a category `other' that denotes expressions/affective states other than the 6 basic ones. The performance measure is the average F1 Score across all 8 categories.
The VA estimation track utilizes 594 videos of around 3M frames of 584 subjects annotated in terms of valence and arousal. The performance measure is the mean Concordance Correlation Coefficient (CCC) of valence and arousal.

\begin{table*}[!t]
\renewcommand\arraystretch{1.05} 
\centering
  \caption{The expression F1 scores~(in \%) of models that are trained and tested on different folds (including the original training/validation set of \textit{Aff-Wild2} dataset).}
  \label{tab:exp_F1_val}
    \vspace{-0.7em}

\setlength{\tabcolsep}{0.7 em}
\renewcommand\arraystretch{1.2}
  \begin{tabular}{l|cccccccc|c}
    \hline
    \textbf{Val Set} &\textbf{Neutral} &\textbf{Anger} &\textbf{Disgust} &\textbf{Fear} &\textbf{Happiness} &\textbf{Sadness} &\textbf{Surprise} &\textbf{Other} &\textbf{Avg.}\\[1pt]
    \hline \hline
    Official&  70.21 & 73.93 & 50.34 & 21.83 & 59.05 & 66.41 & 36.51 & 66.11 & 55.55 \\
    fold-1 &   70.06 & 37.21 & 32.12 & 22.71 & 61.77 & 77.61 & 45.62 & 51.58 & 49.83 \\
    fold-2 &   67.36 & 44.45 & 21.21 & 42.50 & 62.22 & 78.24 & 36.67 & 70.00 & 52.83 \\
    fold-3  &  73.64 & 71.60 & 45.01 & 23.25 & 47.67 & 77.05 & 46.81 & 65.56 & 56.32 \\
    fold-4 &   65.41 & 71.00 & 53.70 & 23.27 & 61.62 & 61.79 & 27.76 & 72.68 & 54.65 \\
    fold-5 &   64.03 & 31.23 & 35.66 & 67.64 & 67.97 & 69.75 & 52.12 & 55.64 & 55.51 \\
    
    \hline
  \end{tabular}
    \label{exp:expr}
\end{table*}

\begin{table*}[!htp]
\centering
\caption{The Pearson’s correlations of models that are trained and tested on different folds (including the original training/validation set of Hume-Vidmimic2.).}
\vspace{-0.7em}

\setlength{\tabcolsep}{0.6 em}
\renewcommand\arraystretch{1.2}
\begin{tabular}{l|cccccc|c} 
\hline 
         & \textbf{Admiration} & \textbf{Amusement} & \textbf{Determination} & \textbf{Empathic Pain} & \textbf{Excitement} & \textbf{Joy}    & \textbf{Avg.}    \\ 
\hline\hline
Official & 0.5942     & 0.4982    & 0.5090        & 0.2275        & 0.4961     & 0.4580 & 0.4638  \\
fold-1   & 0.5880     & 0.4842    & 0.4914        & 0.2089        & 0.4852     & 0.4338 & 0.4486  \\
fold-2   & 0.5193     & 0.4385    & 0.4031        & 0.3715        & 0.3734     & 0.3717 & 0.4129  \\
fold-3   & 0.5195     & 0.4496    & 0.3947        & 0.4924        & 0.4129     & 0.3843 & 0.4422  \\
fold-4   & 0.5955     & 0.4950    & 0.5134        & 0.2492        & 0.5068     & 0.4576 & 0.4696  \\
fold-5   & 0.5199     & 0.4377    & 0.4040        & 0.3739        & 0.3717     & 0.3719 & 0.4132  \\
\hline
\end{tabular}
\label{tab:exp_mimic}

\end{table*}

The fourth track of ABAW6 utilizes 56 videos which are from the C-EXPR-DB database. The complete C-EXPR-DB dataset contains 400 videos totaling approximately 200,000 frames, with each frame annotated for 12 compound expressions. For this track, the task is to predict 7 compound expressions for each frame in a subset of the C-EXPR-DB videos. Specifically, the 7 compound expressions are Fearfully Surprised, Happily Surprised, Sadly Surprised, Disgustedly Surprised, Angrily Surprised, Sadly Fearful, and Sadly Angry. The evaluation metric for this track is the average F1 Score across all 7 categories.

The fifth track of ABAW6 is based on the multimodal Hume-Vidmimic2 dataset which consists of more than 15,000 videos totaling over 25 hours. Each subject of this dataset needs to imitate a `seed' video, replicating the specific emotion displayed by the individual in the video. Then the imitators are asked to annotate the emotional intensity of the seed video using a range of predefined emotional categories~(Admiration, Amusement, Determination, Empathic Pain, Excitement, and Joy). A normalized score from 0 to 1 is provided as a ground truth value for each seed video and each performance video of the imitator. The evaluation metric for this track is the average Pearson's correlation across the 6 emotion dimensions.

In addition to the official datasets mentioned above, we also used some additional data from the open-source and private datasets. For the AU detection track, we use the extra dataset BP4D~\cite{ZhangBP4D} to supplement some of the limited AU categories in Aff-wild2. For the expression recognition track, we use the extra dataset RAF-DB~\cite{li2017reliable} and AffectNet~\cite{mollahosseini2017affectnet} to supplement the Anger, Disgust, and Fear data. For the fourth track, we utilize our private video dataset and annotated these videos based on the rules of 7 compound expressions for training and testing.

\subsection{Implementatal Setting}

We utilize retinaface~\cite{deng2020retinaface} to detect faces for each frame and normalize them to a size of 224x224. We pre-train an MAE on a large facial images dataset that consists of several open-source face images datasets~(i.e., AffectNet~\cite{mollahosseini2017affectnet}, CASIA-WebFace~\cite{CASIA-Webface}, CelebA~\cite{CelebA} and IMDB-WIKI~\cite{IMDB-WIKI}, Webface260M~\cite{zhu2021webface260m}). We use this MAE as the basic feature extractor to capture the visual information for facial images in each track. The pre-training process is trained for 800 epochs with a batch size of 4096 on 8 NVIDIA A30 GPUs, using the AdamW optimizer~\cite{loshchilov2017adamw}. 
For the tasks of AU detection, expression recognition and VA estimation, we incorporate the temporal, audio, and other information to further improve the performance. At this stage, the training data consists of continuous video clips of 100 frames. The learning rate is set as 0.0001 using the AdamW optimizer. To reduce the gap caused by data division, we conduct five-fold cross-validation for all the tracks.

\begin{table}[!t]
\renewcommand\arraystretch{1.05} 
\centering
  \caption{The VA CCC scores of models that are trained and tested on different folds (including the original training/validation set of \textit{Aff-Wild2} dataset).}
  \label{tab:va_F1_val}
\setlength{\tabcolsep}{1.1 em}
\renewcommand\arraystretch{1.0}
  \begin{tabular}{l|cc|c}
    \hline
    \textbf{Val Set} &\textbf{Valence} &\textbf{Arousal} & \textbf{Avg.}\\[1pt]
    \hline \hline
    Official&  0.5523 & 0.6531 &  0.6027\\
    fold-1 &  0.6408  & 0.6195 &  0.6302 \\
    fold-2 &  0.6033  & 0.6758 &  0.6395\\
    fold-3  &  0.6773 & 0.6961 &  0.6867 \\
    fold-4 &   0.6752 & 0.6486 &  0.6619 \\
    fold-5 &   0.6591 & 0.7019 &  0.6801\\
    
    \hline
  \end{tabular}
\end{table}

    

\subsection{Results for AU Detection}
In this section, we show our final results for the task of AU detection. The model is evaluated by the average F1 score for 12 AUs. Table \ref{exp:au} presents the F1 results on the official validation set and five-fold cross-validation set.

\subsection{Results for Expression Recognition}
In this section, we show our final results for the task of expression recognition. The model is evaluated by the average F1 score for 8 categories.  Table ~\ref{tab:exp_F1_val} presents the F1 results on the official validation set and five-fold cross-validation set.

\subsection{Results for VA Estimation}
In this section, we show our final results for the task of VA estimation. The model is evaluated by CCC for valence and arousal.  Table~\ref{tab:va_F1_val} presents the F1 results on the official validation set and five-fold cross-validation set.

\subsection{Results for EMI Estimation}
In this section, we show our final results for the task of EMI estimation. The model is evaluated by Pearson’s correlations. Table~\ref{tab:exp_mimic} presents Pearson’s correlation scores on the official validation set and five-fold cross-validation set.

\section{Conclusion}
\label{sec:conclusion}
In summary, our study contributes to advancing Affective Behavior Analysis, aiming to make technology emotionally intelligent. Through a comprehensive evaluation of the ABAW competition, we address five competitive tracks. Our method designs integrate emotional cues from multi-modal data, ensuring robust expression features. 
We achieve significant performance across all tracks, indicating the effectiveness of our approach. 
These results highlight the potential of our method in enhancing human-machine interactions and technological advancements toward devices understanding and responding to human emotions.

{
    \small
    \bibliographystyle{ieeenat_fullname}
    \bibliography{main}
}


\end{document}